# Marker Track: Accurate Fiducial Marker Tracking for Evaluation of Residual Motions During Breath-Hold Radiotherapy

Aimee Guo[1] and Weihua Mao[2] (Mentor)

[1] The Hockaday School, 11600 Welch Rd, Dallas, Texas, USA

[2] Department of Radiation Oncology and Molecular Radiation Sciences, Johns Hopkins University, Baltimore, Maryland, USA

## ABSTRACT

Fiducial marker positions in projection image of cone-beam computed tomography (CBCT) scans have been studied to evaluate daily residual motion during breath-hold radiation therapy. Fiducial marker migration posed challenges in accurately locating markers, prompting the development of a novel algorithm that reconstructs volumetric probability maps of marker locations from filtered gradient maps of projections. This guides the development of a Python-based algorithm to detect fiducial markers in projection images using Meta AI’s Segment Anything Model 2 (SAM 2). Retrospective data from a pancreatic cancer patient with two fiducial markers were analyzed. The three-dimensional (3D) marker positions from simulation computed tomography (CT) were compared to those reconstructed from CBCT images, revealing a decrease in relative distances between markers over time. Fiducial markers were successfully detected in 2777 out of 2786 projection frames. The average standard deviation of superior-inferior (SI) marker positions was 0.56 mm per breath-hold, with differences in average SI positions between two breath-holds in the same scan reaching up to 5.2 mm, and a gap of up to 7.3 mm between the end of the first and beginning of the second breath-hold. 3D marker positions were calculated using projection positions and confirmed marker migration. This method effectively calculates marker probability volume and enables accurate fiducial marker tracking during treatment without requiring any specialized equipment, additional radiation doses, or manual initialization and labeling. It has significant potential for automatically assessing daily residual motion to adjust planning margins, functioning as an adaptive radiation therapy tool.

## 1. Introduction

Accurate patient positioning and effective management of respiratory motion are critical in pancreatic cancer radiation therapy to ensure precise delivery of prescribed doses to tumor targets while minimizing toxicity to adjacent organs at risk (Chang et al, 2009). Deep inspiration breath-hold (DIBH) techniques have gained widespread acceptance as a strategy to mitigate respiratory motion during treatment simulation and cone-beam computed tomography (CBCT) acquisitions. These approaches enhance patient stability, allowing for better alignment during verification and treatment delivery. Nonetheless, several studies have reported the presence of residual motion even during breath-hold procedures, which can undermine treatment accuracy. A variety of investigations have been conducted to quantify this residual motion. Blessing et al. employed manual determination of diaphragm dome positions on CBCT projections to assess residual motion while utilizing an active breath coordinator (ABC) DIBH (Blessing et al, 2018). Ultrasound techniques were used to detect diaphragmatic dome positions, comparing the findings with those obtained from manual assessments in CBCT (Boda-Heggemann et al, 2019; Vogel et al, 2018). Implanted fiducial markers were tracked by positioning kilovoltage radiographs or intermittent kV imaging (Dawson et al, 2001; Zeng et al, 2021; Zeng et al, 2019). While these approaches have advanced our understanding of motion management, they often rely on manual techniques that can be time-consuming. To solve this problem, we developed an automatic algorithm to track fiducial markers on CBCT projections (Mao et al, 2024). This algorithm generates digitally reconstructed radiograph (DRR) of markers from initial simulation computed tomography (CT) to match markers on each projection. However, implanted fiducial markers might migrate or deform differently in soft tissues, leading to possible mismatch of markers between DRR and projections. This paper proposes a novel method to reconstruct daily approximate 3D positions of markers to guide accurate tracking of markers on each projection. In addition, Meta AI's newly published Segment Anything Model 2 (SAM 2) (Ravi et al, 2024) was an innovative image and video segmentation method that makes segmentation easier and more efficient with its exceptional accuracy. SAM 2 enabled precise segmentation of any object in both videos and images, making it a valuable ally in our task of tracking marker positions throughout the continuous capturing of CBCT images.

## 2. Materials and Methods

### 2.1 Patient Data

A pancreatic cancer patient treated in April 2024 was enrolled in this retrospective study under an approved institutional review board protocol (IRB00395200). Two fiducial markers were endoscopically implanted in the pancreas. These LumiCoil STRAIGHT platinum fiducial markers (Boston Scientific, Marlborough, MA, USA) had dimensions of 0.48 mm in diameter and 5 mm in length. To ensure consistent positioning, the patient was immobilized using a Vac-Lok (CIVCO Medical Solutions, Coralville, IA, USA) along with a Wing Board (CIVCO Medical Solutions). DIBH was facilitated by an active breathing coordinator (ABC, Elekta, Stockholm, Sweden), the patient trained to follow a breathing protocol involving inhalation, exhalation, and breath-hold. This patient underwent four contrast-enhanced CT simulation scans during separate DIBH sessions using the Brilliance Big Bore CT simulator (Philips Medical Systems, Cleveland, OH, USA). One set of CT images was selected for contouring and treatment planning in RayStation (RaySearch Laboratories AB, Stockholm, Sweden). The CT images had a slice thickness of 2 mm and a planar pixel size of approximately 1.4 mm. Fiducial markers were contoured using a bone window with a Hounsfield Unit range of [450, 1600]. Stereotactic body radiation therapy (40 Gy in 5 fractions) was planned with volumetric modulated arc therapy (VMAT) treatment with two arcs. Treatments were administered using a VersaHD linear accelerators (Elekta, Stockholm, Sweden). A partial arc (200-degree) rotation cone-beam CT (CBCT) was performed with acquisition settings of 120 kV, 40 mA, and 40 ms. The source-to-axis distance (SAD) was 1000 mm, and the source-to-imager distance (SID) was 1536 mm, with a projection image resolution of 512 × 512 pixels and a pixel size of 0.8 mm. The maximum breath-hold (BH) duration was 25 seconds, and CBCT scans were acquired over two consecutive breath-holds. Reconstructed CBCT images were rigidly registered to the planning CT based on bony landmarks, with final adjustments made using the fiducial markers. Following multiple reviews, six-dimensional corrections were sent to the Hexapod (Elekta) to adjust the treatment couch. Verification CBCT scans were obtained immediately before and/or during the middle of treatment, with beams delivered under the same DIBH conditions. However, it was noted that marker positions in CBCT were very different from those in the simulation CT in the first two fractions of treatment. Four additional simulation CT scans were acquired, and a new plan was generated with a prescription of 36 Gy in 8 fractions. A total of 14 CBCT scans were acquired in 12 days during the course of treatment.

## 2.2 Data Analysis

To facilitate further analysis, planning CT images, reconstructed CBCT images, CBCT projection data, and ABC results were de-identified and exported from clinical systems. Analyses were conducted using in-house MATLAB (MathWorks, Natick, Massachusetts) and Python (Python.org) software. Fiducial mass centers were computed within fiducial structures to determine their initial positions in the planning CT. Fiducial positions were determined in all CT and CBCT images using MATLAB program and compared across the simulation CT scans by registering the planning CT to the other CT sets, focusing on the markers. Distances between fiducial markers were calculated across all four sets of CT images to evaluate potential variations. Further studies on detecting markers on CBCT projections were conducted as described below.

### 2.2.1 Updating Expected Markers' 3D Positions

For all $M$ fiducial markers, their original positions $m_k, k \in [1, M]$ obtained during the treatment planning stage might become inaccurate in daily treatment process. In this first step, we proposed a new method to update the expected marker positions based on different frames of the captured CBCT images in daily treatment.

The 3D coordinates of original marker positions were averaged to determine their center $c = \frac{\sum_k m_k}{M}$. Then, a bounding box that enclosed the original marker positions was computed. The longest length of the box's three dimensions was used to create a cube centered at $\boldsymbol{c}$. Finally, the cube was subdivided into $50 \times 50 \times 50$ voxels. In later steps, we would acquire each voxel's probability of containing a marker based on the information from different frames of captured CBCT images.

For each 2D CBCT projection $i$, a noise-suppressed gradient image $g_i$ was calculated (Figure 1) (Mao et al, 2024), where the partial derivatives approximated by the difference of values between neighboring pixels were screened by a preset threshold $\mu = 32$. That is, if the partial derivative was smaller than $\mu$, we set it as zero. Since the marker's density was the highest in the image, this noise suppression strategy effectively removes majority of unnecessary information from the image. Afterwards, the gradient images were normalized into grayscale 2D arrays with values between 0 and 1, representing the probability $p_i$ of being a marker in the captured 2D projection: $p_i = normalize(g_i)$.

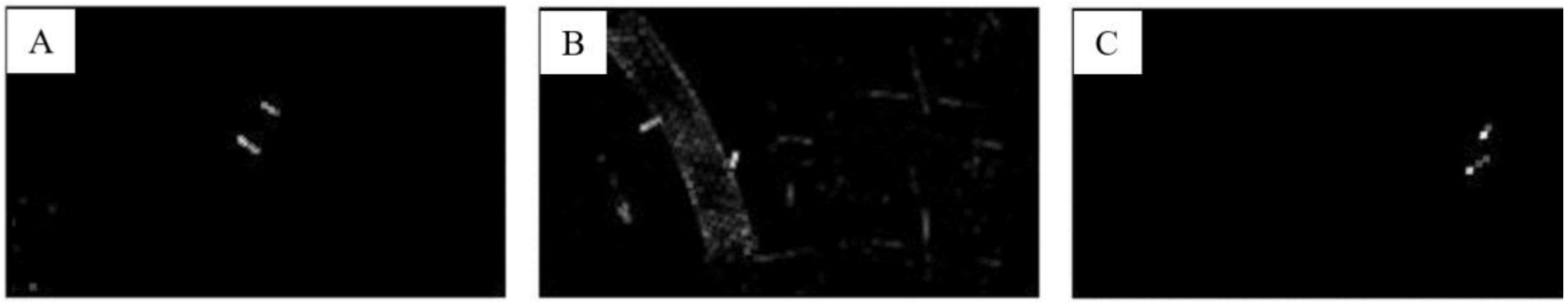

Figure 1. Noise-suppressed gradient images. (A)-(C) are results of three projections.

The 3D coordinates $(x, y, z)$ of each voxel $\mathbf{v}$ were projected into 2D coordinates $(u, v)$ shown in Figure 2. Their connections were presented in Equations (1) and (2) (Mao et al, 2008; Mao et al, 2024).

$$u = \frac{\mathrm{SID} \cdot (x \cdot \cos\phi - z \cdot \sin\phi)}{\mathrm{SAD} - x \cdot \sin\phi - z \cdot \cos\phi} \tag{1}$$

$$v = \frac{\mathrm{SID} \cdot y}{\mathrm{SAD} - x \cdot \sin\phi - z \cdot \cos\phi} \tag{2}$$

Here, SID is the source-to-image distance (1,536 mm), SAD is source-to-axis distance (1,000 mm), and $\phi$ is the gantry angle of the captured CBCT projection.

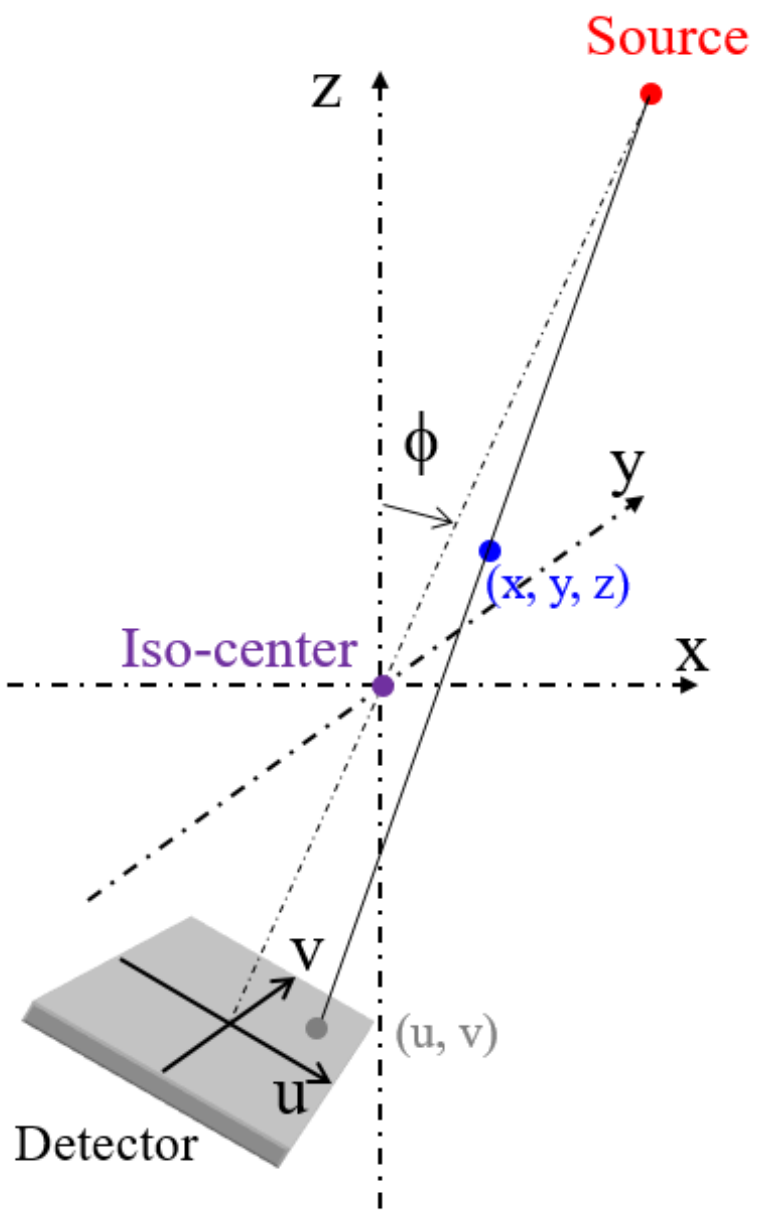

Figure 2. Schematic diagram of the coordinate system.

Next, all probability values $p_i(\pi_i(\mathbf{v}))$ were summed up and stored into each respective voxel to represent the probability of each voxel being the marker: $f(\mathbf{v}) = \sum_i p_i(\pi_i(\mathbf{v}))$ as shown in Figure 3. Here $\pi_i(\mathbf{v})$ was the projection of voxel $\mathbf{v}$ as described in Equations (1) and (2). We call such voxelized representation of $f(\mathbf{v})$ as *marker probability volume*.

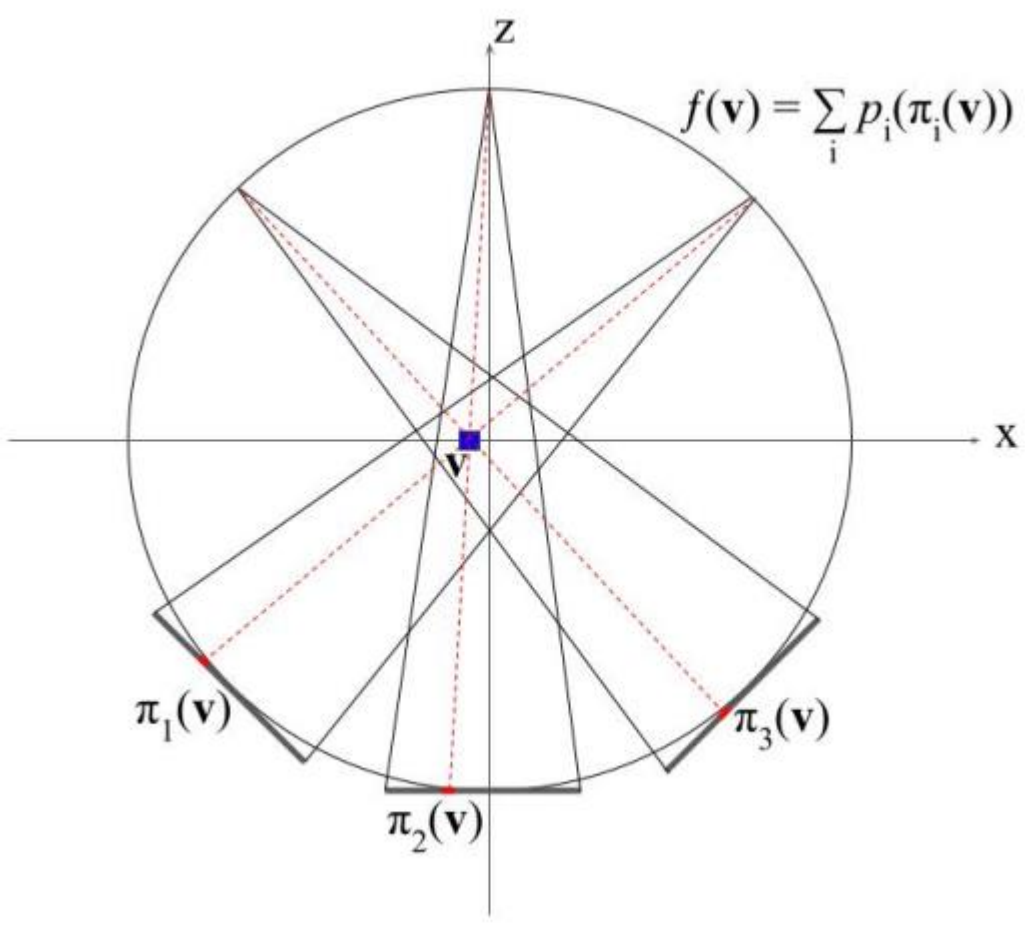

Figure 3. Computing a voxel's marker probability $f(\mathbf{v})$ by summing up all probability values of its projections $\pi_i(\mathbf{v})$ captured in the marker probability $p_i$ of those 2D CBCT images.

The marker probability volume $f(\mathbf{v})$ computed with this method might contain excess noise around the true marker positions because of the non-zero gradient values of bones and tissues in the calculated two-dimensional (2D) probability images. Excess noise was removed by using a threshold $\lambda$ of 70% of the maximum probability value in the volume (Figure 4). If a voxel's probability value was smaller than the threshold $\lambda$, its value was changed to zero.

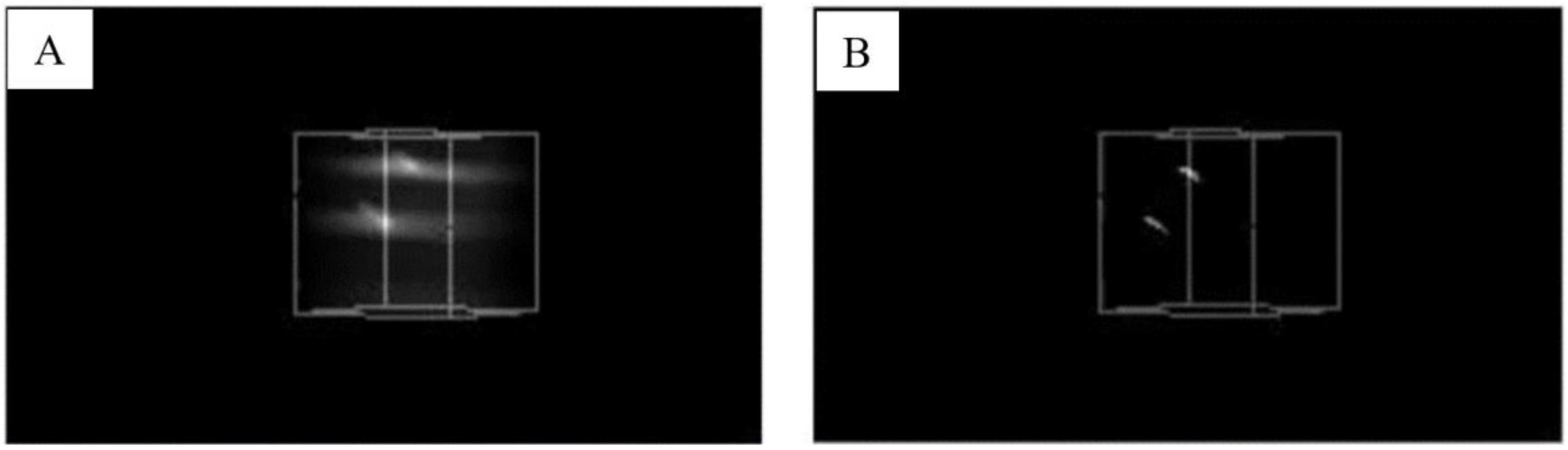

Figure 4. Removing noise in the marker probability volume by thresholding. (A) and (B) are views of probability volume before and after removing noise.

After removing noise in the marker probability volume $f(\mathbf{v})$, grouped three dimensional (3D) clusters were found using flood-fill algorithm implemented with a depth-first search. Each 3D cluster of voxels represents the spatial occupancy of one marker. Then, each cluster's 3D centroid was calculated by averaging all x, y, and z coordinates of voxels in that cluster, representing the corresponding marker's center position.

Each extracted centroid of a cluster represents a marker found in the marker probability volume $f(\mathbf{v})$. If there were more than one marker in the patient body, we need to match each extracted centroid to each expected marker. The 3D centroids were matched to the expected 3D fiducial marker positions by enumerating all possible pairings of 3D centroid to expected marker and finding the pairing with shortest distances between them. With this method, we obtained matched centroids, otherwise known as refined 3D marker positions $\overline{m}_k$.

### 2.2.2 Continuous Tracking of Marker Positions in CBCT Projections

Each refined 3D marker position $\overline{m}_k$ was different for each breath hold, as calculations for each breath hold could be found separately. However, the refined 3D marker positions $\overline{m}_k$ were calculated based on the assumption that the marker was static throughout the imaging process. Although they were very close on each image, there might be some small discrepancies between the refined marker positions $\overline{m}_k$ and the actual positions captured in each projection image. To account for possible marker movement through the imaging process, we propose the following solution to track the real marker positions in each 2D projection image, basing it on the refined 3D marker positions $\overline{m}_k$ obtained above.

For each 2D CBCT projection $i$, we utilized the previous noise-suppressed gradient image $g_i$ by converting it to a regular gradient image $\bar{g}_i$ without a preset threshold, avoiding any potential cut-off of true marker pixels in the CBCT projections. SAM 2 can continuously and accurately segment a video with excess noise. The regular gradient image $\bar{g}_i$ is inputted into the model as shown in Figure 5(A) and is fed into SAM 2's image encoder (Figure 5(B)). Subsequently, memory attention (shown in Figure 5(C)) is used to recall features and predictions of previous video frames ($\bar{g}_j, j < i$) to segment new video frames ($\bar{g}_i$), and it is constantly updated with new segmentations as shown in Figure 5(G, H, I). Point prompts (shown as green stars in Figure 5(D)) are obtained from previously calculated refined 3D marker positions $\overline{m}_k$ and are provided to the prompt encoder (Figure 5(E)). SAM 2's prompt encoder accepts point prompts, bounding boxes, or mask prompts, and generated frame embeddings are inputted with memory attention into the mask decoder (Figure 5(F)), which generates a final segmentation for each frame (shown in Figure 5(G)).

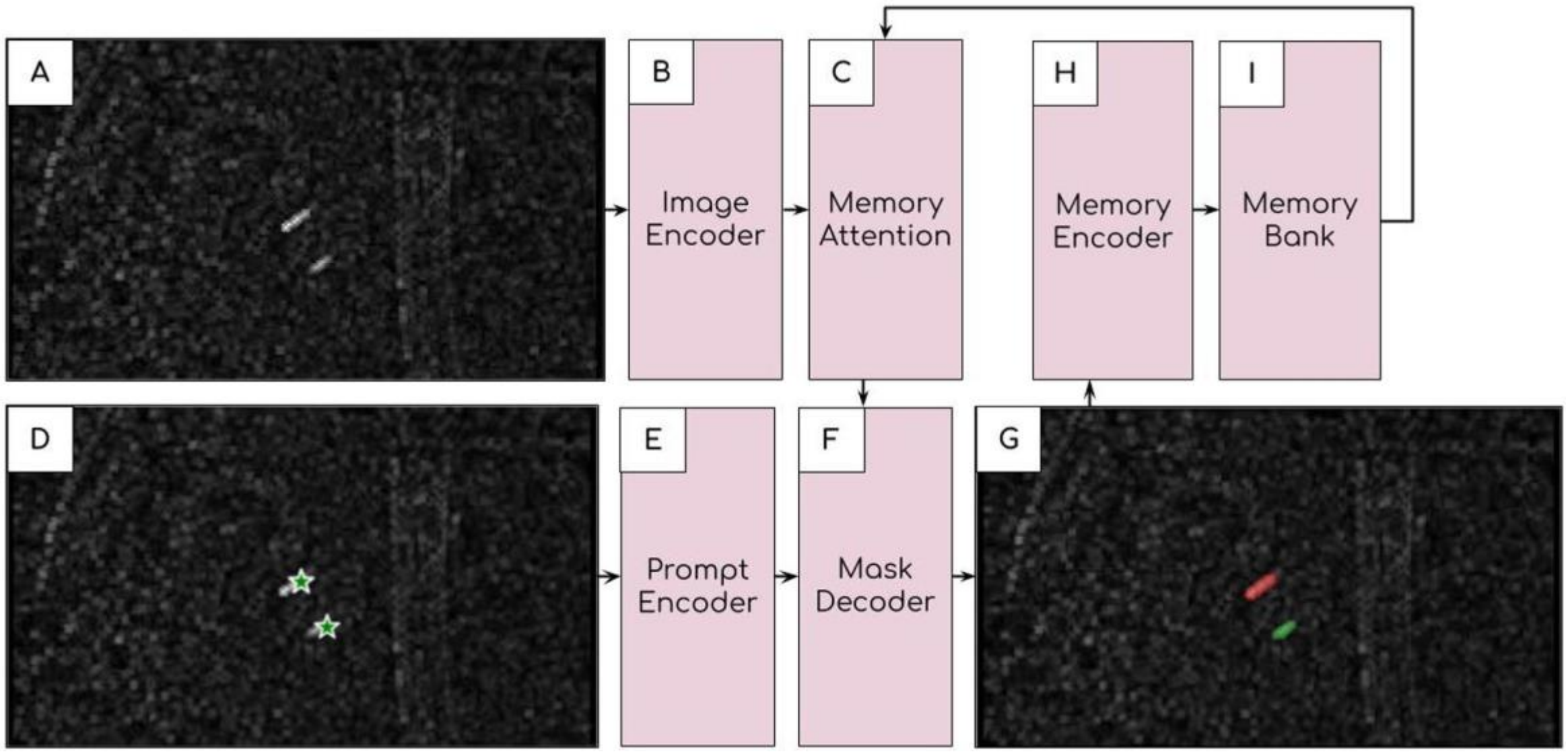

Figure 5. SAM 2 model architecture and our processing pipeline. (A) input image; (B) fed into SAM 2's image encoder; (C) memory attention to recall features and predictions of previous video frames; (D) point prompts (using refined expected marker positions); (E) prompt encoder accepts point prompts, bounding boxes, or mask prompts, and generates frame embeddings; (F) input in mask decoder; (G) generates a segmentation; (H) & (I) segmentation results inputted into memory encoder and memory bank and are sent back to memory attention.

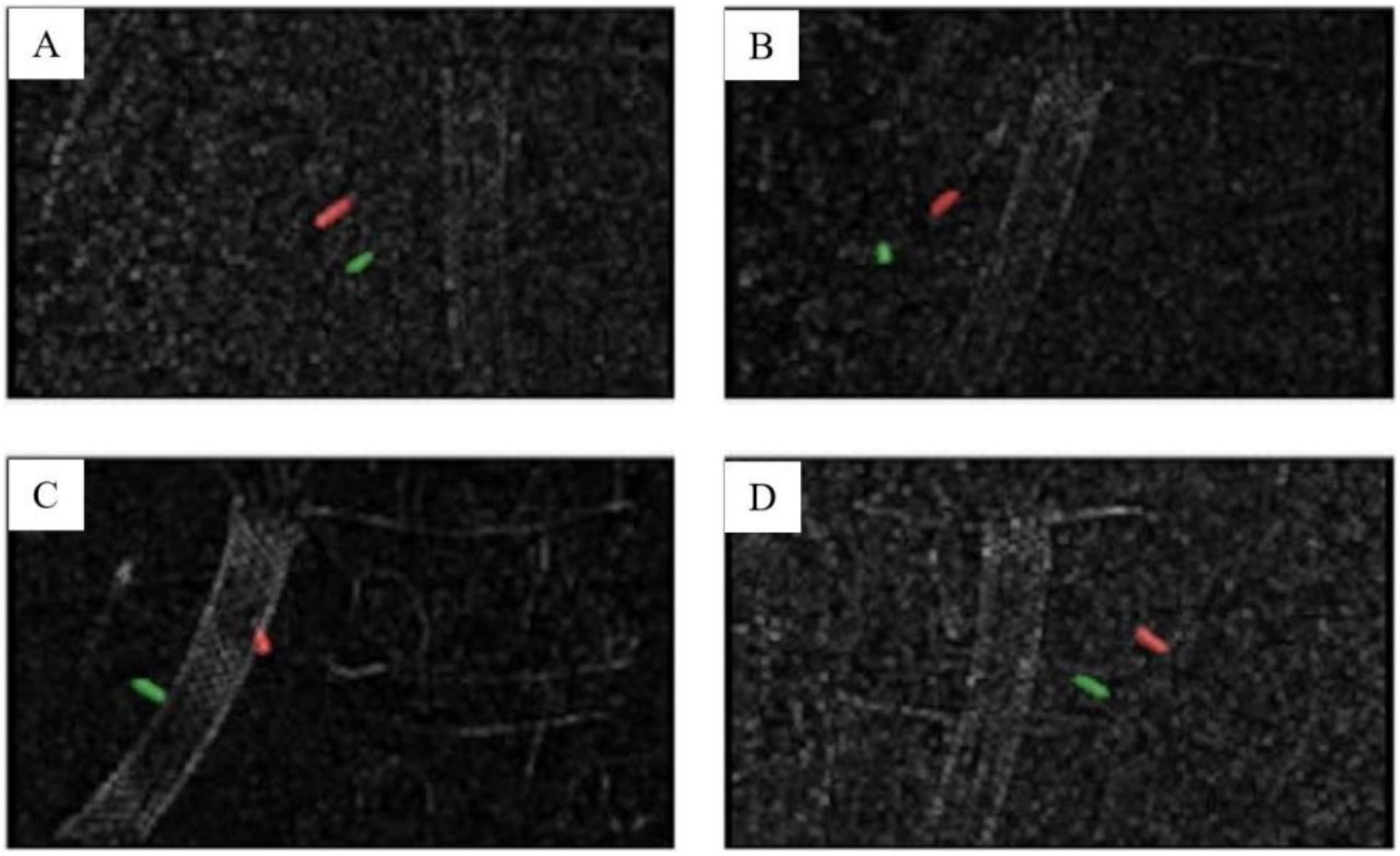

Figure 6. Continuous tracking of two markers (red and green) from the regular gradient images (without noise-suppression) of CBCT projections. (A)-(D) are results of different projections.

After using SAM 2 and retrieving the segmentations (Figure 6) of all markers in each CBCT projection, the centers of the markers were calculated by averaging all pixel coordinates in the segmented marker mask.

### 2.2.3 Postprocessing

A screening analysis was conducted on the 2D fiducial marker positions to identify and remove outliers based on expected position trendlines. After filtering, three-dimensional (3D) positions were calculated from the projection data. The lateral (X) and vertical (Z) coordinates of each fiducial marker were determined using the lateral projection positions (u) from each breath-hold, as described by Equation (1). These coordinates were then used to calculate expected projection positions. Any projection showing a lateral position deviating by more than 5 mm from the expected value was considered an outlier and excluded from further analysis.

The superior-inferior (SI) or longitudinal (Y) coordinates of each fiducial marker were calculated for each projection according to Equation (2) using lateral and vertical coordinate results per breath-hold. Screening analyses were performed separately for each breath-hold session. A third-order polynomial curve fitting was applied to the longitudinal coordinates to model the expected positions. Projections with detected longitudinal positions deviating by more than 3 mm from the curve-fit results were classified as outliers and excluded from further analysis.

The SI coordinates were analyzed separately for each breath-hold session as well as for the entire scan, which included both breath-holds. These included the average positions, maximum deviations from the average, and standard deviations. The difference in average positions between the two breath-holds in the same scan was also calculated. Additionally, the positional gap - defined as the difference between the end of the first breath-hold and the beginning of the second - was determined for each scan.

## 3. Results

Both markers were detected in 8 sets of CT and 14 sets of CBCT images. The relative distances in three directions and absolute distance are illustrated in Figure 7.

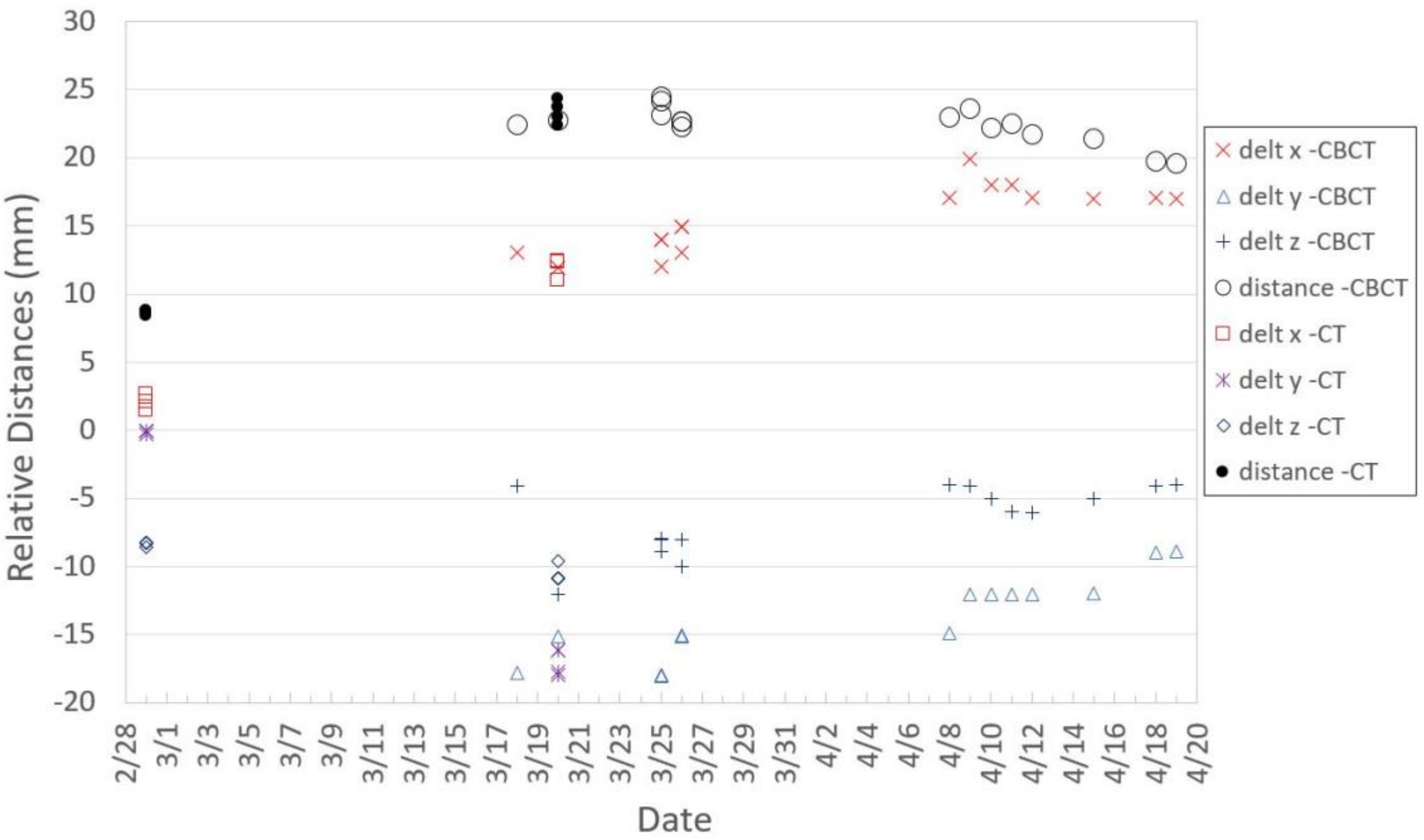

Figure 7. Relative distance between two markers in three directions and total distance found in CT and CBCT images.

In total, 2,786 frames of projections were analyzed and both markers were successfully detected in 2,777 frames. Figure 8 shows detected results during one CBCT scan. SI positions of both markers in all projections were analyzed and are listed in Table 1. The maximum standard deviation was 1.1 mm per breath-hold and 2.6 mm per scan. The difference of average positions between two breath-holds in the same scan was up to 5.2 mm, as shown in Figure 8. The maximum gap between the end of the first breath-hold and the beginning of the second breath-hold was up to 7.3 mm.

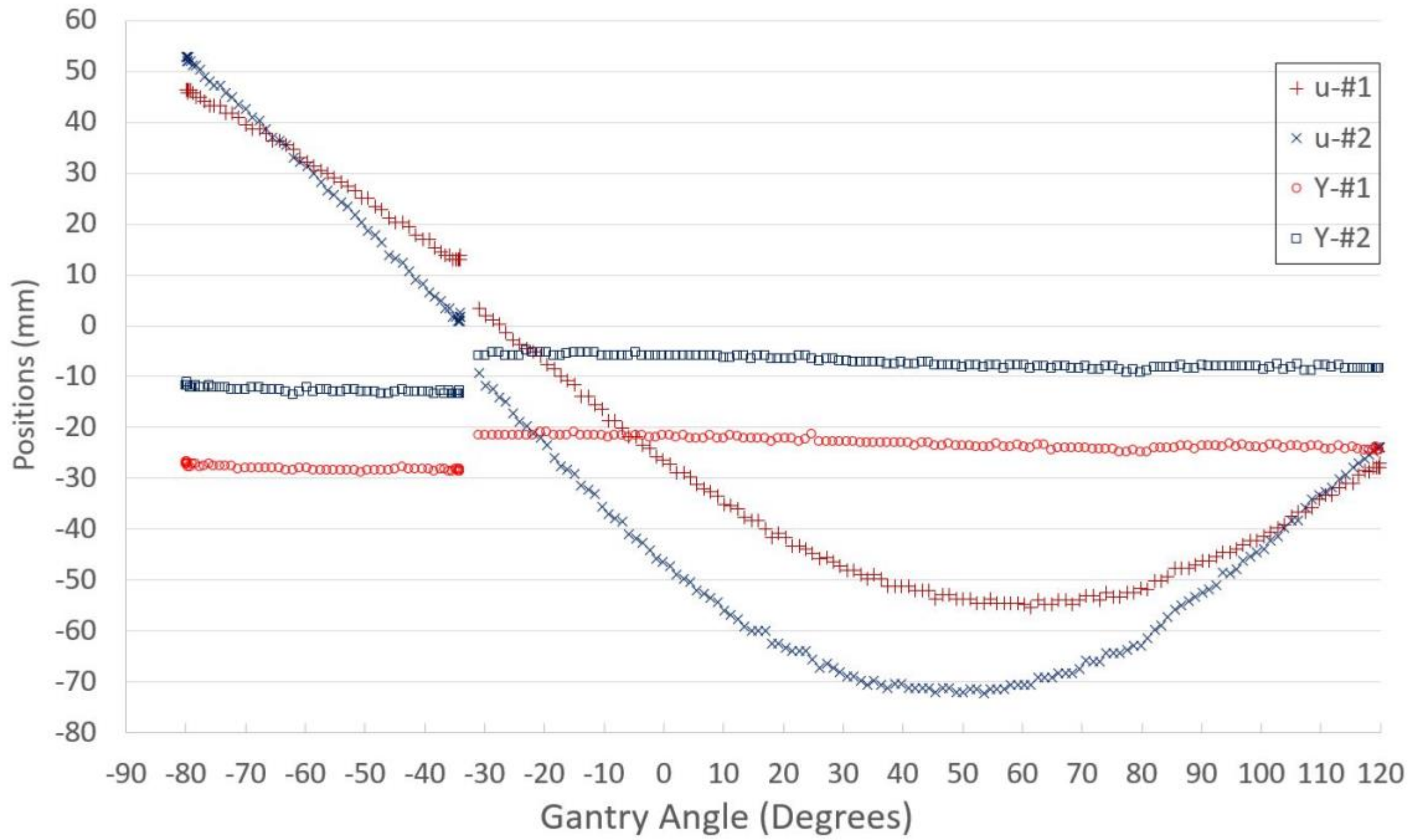

Figure 8. Detected marker lateral positions (u) on imager and SI positions (Y) of both markers.

3D positions were calculated for each breath-hold and both breath-holds, respectively. The relative distances between markers were calculated per scan and are displayed in Figure 9.

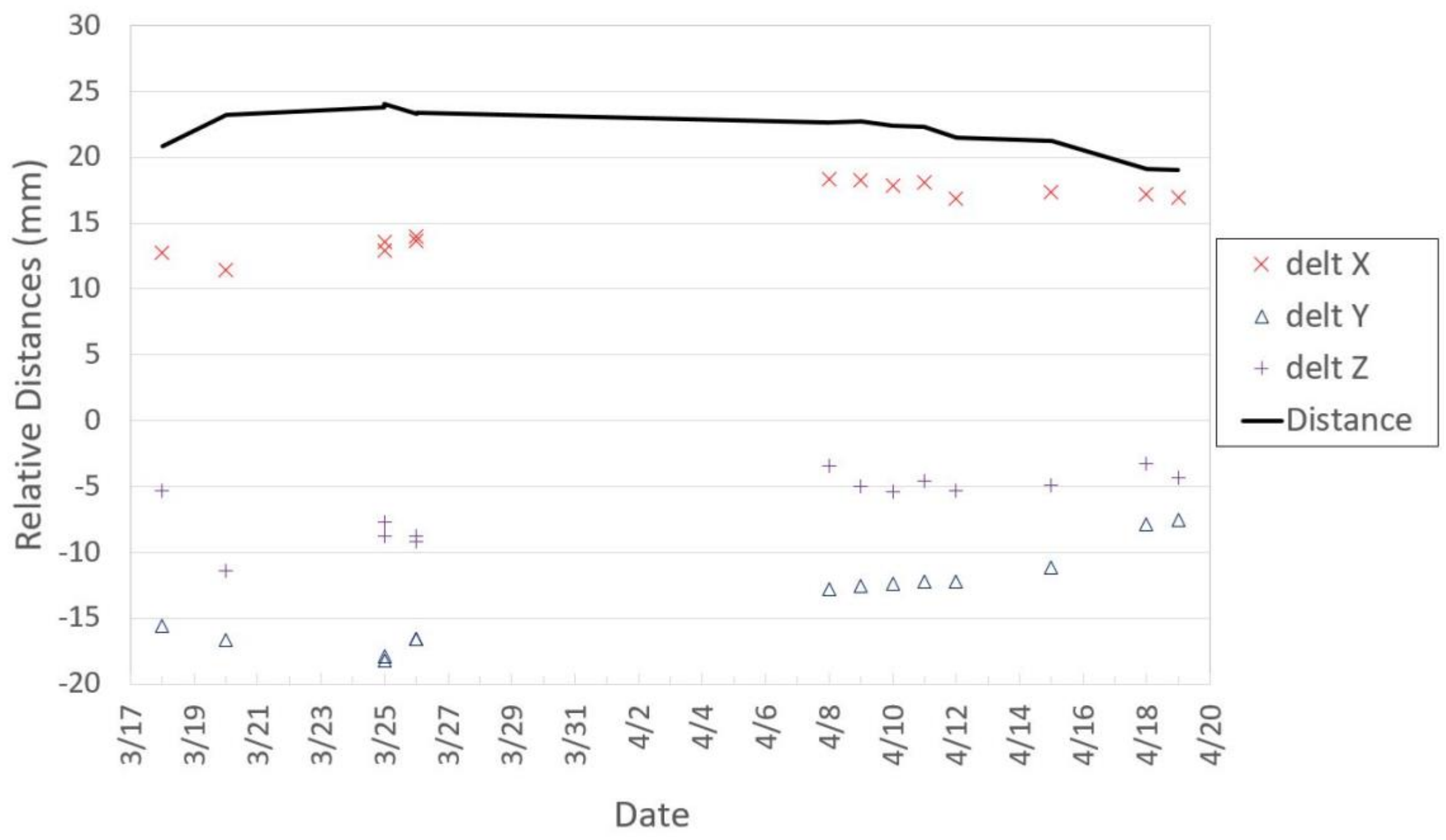

Figure 9. Relative distance between two markers in three directions and total distance found in CBCT images.

Table 1. Average SI positions of both markers detected in 14 CBCT scans. Maximum deviation (Max Dev) and standard deviation (Std Dev) of marker positions were listed for each breath-hold and both breath-holds, respectively. Differences (Avg Diff) between average positions of two breath-holds and the Gap (the difference between end position of the 1st breath-hold and beginning position of the 2nd breath-hold) are listed (Unit: mm).

| Date | 1st breath-hold | | 2nd breath-hold | | both breath-holds | | Avg Diff | Gap |
|---|---|---|---|---|---|---|---|---|
| | Max Dev | Std Dev | Max Dev | Std Dev | Max Dev | Std Dev | | |
| 3/18 | 2.0 | 1.1 | 1.3 | 0.5 | 4.3 | 2.6 | 5.2 | 7.3 |
| 3/20 | 1.0 | 0.4 | 1.3 | 0.7 | 2.3 | 0.8 | 1.4 | 2.1 |
| 3/25 | 1.1 | 0.4 | 0.4 | 0.1 | 2.3 | 1.3 | 2.9 | 3.3 |
| 3/25 | 1.5 | 0.6 | 1.5 | 0.9 | 2.7 | 1.2 | 2.2 | 1.9 |
| 3/26 | 1.4 | 0.5 | 1.2 | 0.7 | 1.6 | 0.6 | 0.6 | 0.2 |
| 3/26 | 1.3 | 0.7 | 1.5 | 0.8 | 1.6 | 0.7 | 0.2 | 0.4 |
| 4/8 | 0.9 | 0.4 | 1.3 | 0.7 | 1.2 | 0.6 | 0.2 | 0.8 |
| 4/9 | 1.4 | 0.6 | 1.5 | 0.7 | 1.6 | 0.7 | 0.8 | 0.6 |
| 4/10 | 0.8 | 0.3 | 1.7 | 0.9 | 2.0 | 0.7 | 0.6 | 1.3 |
| 4/11 | 0.7 | 0.2 | 2.1 | 1.1 | 2.0 | 0.8 | 0.2 | 1.3 |
| 4/12 | 0.6 | 0.2 | 1.1 | 0.5 | 3.3 | 1.4 | 3.1 | 2.6 |
| 4/15 | 1.1 | 0.3 | 0.8 | 0.3 | 4.0 | 2.3 | 4.8 | 4.1 |
| 4/18 | 1.0 | 0.4 | 0.9 | 0.5 | 2.1 | 1.0 | 1.9 | 3.6 |
| 4/19 | 1.0 | 0.5 | 1.0 | 0.5 | 1.9 | 0.7 | 1.1 | 0.9 |

## 4. Discussions

This method reconstructs 3D probability distributions of fiducial markers, guiding the determination of accurate marker positions in projections. The standard deviation can be regarded as an indication of positional variation. The average standard deviation per breath-hold is 0.5 mm, meaning that approximately 95% of the positions fall within a 1 mm range. It should be noted that the fiducial markers were 5 mm in length, with a pixel size of approximately 0.5 mm when projected to the isocenter. The 3D positions obtained from projection data confirm marker migration, consistent with the results from 3D images. This verifies that the method accurately determines the marker center.

Our results indicate that the average standard deviation of positions per scan was 1.1 mm ranging from 0.6 to 2.6 mm. A planning margin of 4 mm was used to expand the clinical target volume for initial and re-planning of this case. This means that, on average, 99.98% of the positions fell within the range.  There were two scans with standard deviations of 2.3 and 2.6 mm, which indicated 95.9% and 87.9% coverage of motions.  This method enables detailed and accurate residual motion detection during breath-hold, without requiring any specialized equipment, additional radiation doses, or manual initialization and labeling. It has significant potential for automatically assessing daily residual motion to adjust planning margins, functioning as an adaptive radiation therapy tool.

This approach differs from conventional CBCT volumetric image reconstruction. It utilizes filtered gradient maps from the projections, specifically focusing on the markers, without requiring a large number of projections over a wide range of gantry angles. For the first breath-hold, gantry angles ranged from 98 to 165 degrees, and for the second breath-hold, between 32 and 101 degrees. As a result, this method effectively eliminates noise. In this case, the patient had a metal stent implanted near the target, which produces a much stronger signal than the markers in the CBCT images, but it was eliminated from the probability distributions by the bounding box and the filtered gradient maps. This makes marker detection more straightforward in the probability maps than in the CBCT images.

Utilizing Meta AI's SAM 2 segmentation model, our method helps users accurately and efficiently track and segment objects in videos. Previous Meta AI segmentation models, such as SAM (Kirillov et al, 2023), which can only segment single images, have become widely used in biomedical engineering due to their precision, efficiency, and flexibility. Used in various imaging modalities like magnetic resonance imaging (MRI) and CT scans, and incorporated into new segmentation models like MedSAM (Ma et al, 2024), SAM has proven to be a valuable tool for medical professionals. As a deep learning model, SAM supports multiple user inputs, demonstrating its adaptability to various data, which is particularly useful for doctors working with diverse or uncommon datasets. In contrast to Meta AI's SAM, SAM 2 is designed for temporal sequencing, using a memory system to reference previous video frames during segmentation. This feature enables easy object tracking in videos and can be applied to our tracking of fiducial markers in scans.

Additionally, this method could be applied to free-breathing CBCT projections. In theory, two projections from different directions should be sufficient to determine the 3D positions of static markers. A series of projections could provide probabilistic estimates of 3D marker positions, which would be especially useful for constraining the range of moving markers.

However, our algorithm has some limitations for unusual projection images with high noise levels. For these images, noise suppression is insufficient to make the markers visually distinct, which results in a reconstructed 3D marker probability volume that still contains significant noise. This can complicate the subsequent tracking process. Although these cases are rare in our experiments—our captured CBCT dataset containing only one such case—they could potentially be addressed in the future with more advanced noise-suppression algorithms.

## 5. Conclusions

A new algorithm has been developed to reconstruct marker probability volumes and enable accurate tracking of fiducial markers in CBCT projections. This will be very helpful for evaluating daily residual motion during breath-hold and motion patterns in challenging cases to assess planning margins.